\scrollmode
\UseRawInputEncoding
\documentclass[final]{cvpr}
\usepackage{lineno,hyperref}
\usepackage{amsmath}
\usepackage{times}
\usepackage{epsfig}
\usepackage{graphicx}
\usepackage{amssymb}

\usepackage{bm}
\usepackage{array}
\usepackage{multirow}
\usepackage{color}
\usepackage{fonttable}

\def\cvprPaperID{5941} 
\def\confYear{CVPR 2021}
\makeatletter
\newcommand{\thickhline}{%
    \noalign {\ifnum 0=`}\fi \hrule height 1.1pt
    \futurelet \reserved@a \@xhline
}

\begin{document}

\title{Context Sensing Attention Network for Video-based Person Re-identification}

\author{
Kan Wang$^{1,2}$
\quad Changxing Ding$^{3}$ \thanks{Corresponding author}
\quad Jianxin Pang$^{1}$
\quad Xiangmin Xu$^{3}$\\
$^1$ UBTECH Research, UBTECH Robotics, Inc.\quad
$^2$ Shenzhen Institutes of Advanced Technology \\
$^3$ South China University of Technology  \quad \\
{\tt\small kan.wang@ubtrobot.com, chxding@scut.edu.cn, walton@ubtrobot.com, xmxu@scut.edu.cn}
}

\maketitle

\pagestyle{empty}
\thispagestyle{empty}

\begin{abstract}
Video-based person re-identification (ReID) is challenging due to the presence of various interferences in video frames. Recent approaches handle this problem using temporal aggregation strategies. In this work, we propose a novel Context Sensing Attention Network (CSA-Net), which improves both the frame feature extraction and temporal aggregation steps. First, we introduce the Context Sensing Channel Attention (CSCA) module, which emphasizes responses from informative channels for each frame. These informative channels are identified with reference not only to each individual frame, but also to the content of the entire sequence. Therefore, CSCA explores both the individuality of each frame and the global context of the sequence. Second, we propose the Contrastive Feature Aggregation (CFA) module, which predicts frame weights for temporal aggregation. Here, the weight for each frame is determined in a contrastive manner: i.e., not only by the quality of each individual frame, but also by the average quality of the other frames in a sequence. Therefore, it effectively promotes the contribution of relatively good frames. Extensive experimental results on four datasets show that CSA-Net consistently achieves state-of-the-art performance.
\end{abstract}

\section{Introduction}
\begin{figure*}[t]
    \centerline{\includegraphics[width=0.90\textwidth]{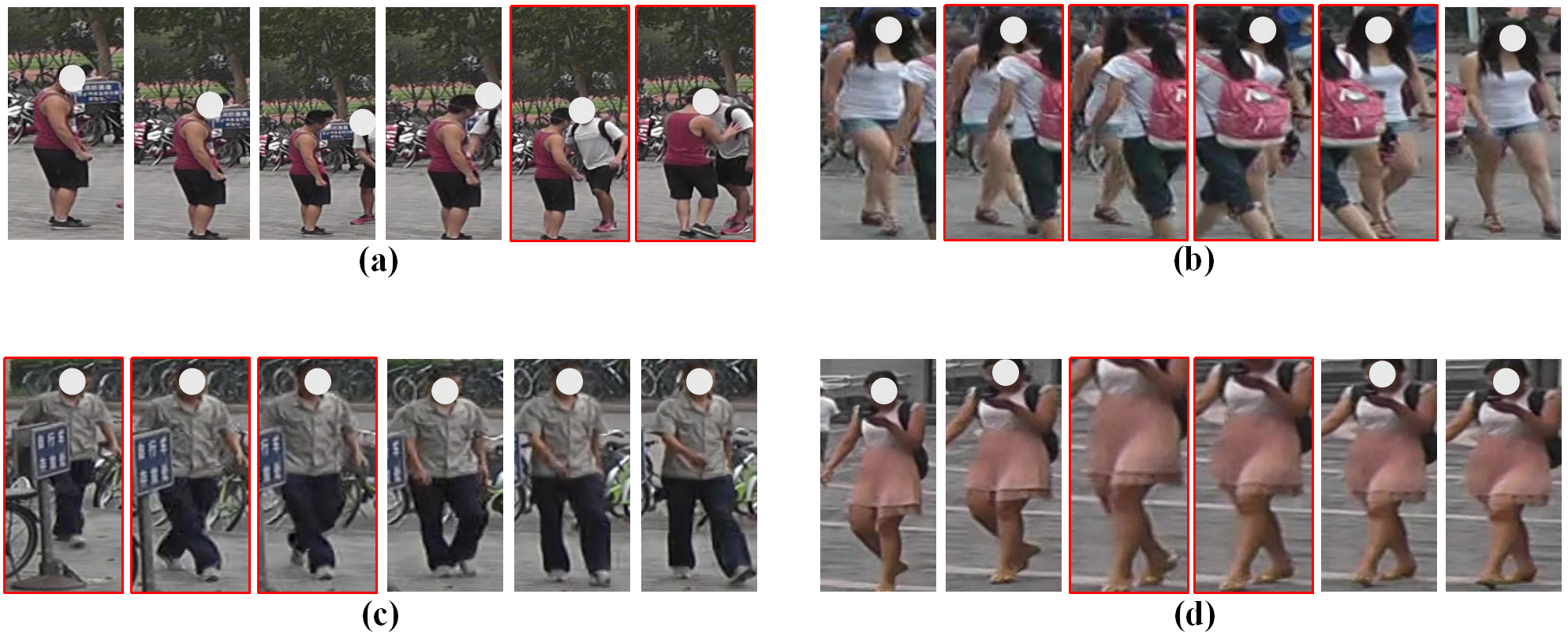}}
     \caption{Examples that illustrate the challenges for video-based ReID. As shown by the images framed in red rectangles, pedestrian appearance is affected by (a-b) interference from other pedestrians, (c) occlusion, and (d) pedestrian detection errors.}
    \label{problem}
    \end{figure*}

  The goal of video-based person re-identification (ReID) is to identify a person of interest using video sequences captured across disjoint camera views \cite{ge2021cross, liu2021watching, zhang2021pixel, zhang2021hat, tang2022harmonious, fang2021set, shen2019multi, yang2017enhancing, fan2018unsupervised, xu2022bire, gao2021mso, wang2021batch}. Compared with individual images, video sequences provide richer cues about pedestrians' identity; therefore, video-based ReID has become an important topic with the widespread usage of surveillance networks \cite{zhong2021glance, ruan2020correlation, pang2022fully, ECCV2016mars, ECCV2020Exploiting, liu2019dense, liu2021viewing, Trans2018pedestrian,TIP2020limulti}. However, as illustrated in Figure \ref{problem}, it remains a challenging problem due to the presence of interference from other pedestrians, occlusion, and pedestrian detection errors.

    A pipeline for video-based ReID typically comprises two sequential steps: \emph{frame feature extraction} and \emph{temporal aggregation}. The first of these steps extracts features from each individual frame, and the second one aggregates these features into a video feature. Existing approaches typically focus on the second step \cite{CVPR2017quality, ICCVChen2017joint,CVPR2020STG,ICCVLi2019Global,CVPR2020MGH,CVPR2020MGRAFA}, and are largely temporal pooling-based. Temporal pooling-based methods predict an attention score for each frame in order to promote the contribution of high-quality frames and suppress that of poor-quality ones. However, these methods usually estimate the score based on the content of each frame alone, often ignoring the temporal context \cite{CVPR2017quality, ICCVChen2017joint}. Subsequent approaches have improved frame features by means of information propagation between frames \cite{CVPR2020MGH, CVPR2020STG, ICCVLi2019Global}. However, information propagation tends to occur between semantically similar frames, meaning that poor-quality frame features may be revised by other poor-quality ones. Moreover, the above works may underestimate the importance of the frame feature extraction step.

    In this work, we demonstrate the advantages of improving both the \emph{frame feature extraction} and \emph{temporal aggregation} steps. Our key observation is that poor-quality frames usually account for only a small proportion of a video sequence, meaning the content in video sequence is stable. Therefore, the temporal context is consistent across most frames. Accordingly, we propose the Context Sensing Attention Network (CSA-Net), which constructs two novel modules that respectively extract discriminative and robust frame- and video-level representations, by exploiting the temporal context of video sequence.


    First, we propose the Context Sensing Channel Attention (CSCA) module, which extracts robust frame-level features. Existing channel attention modules, such as the squeeze-and-excitation (SE) module \cite{cvprse2018}, aim to emphasize responses from informative channels and suppress responses from less useful ones. However, without temporal context, such a module may not correctly infer which channels are informative. For example, as illustrated in Figure \ref{problem}(a), the body region of the dominant pedestrian in some frames is vague when interference between pedestrians occurs; it becomes clear only when we consider the temporal context information of the entire video sequence. CSCA handles the above problem by modulating the responses of the hidden layer in the SE module according to the overall content of one video sequence. In this way, macro-visual patterns that are irrelevant to the dominant pedestrian are suppressed. Moreover, the responses of the CSCA output layer are free from direct modulation, which enables the individuality associated with each single frame to be taken into account. This is because micro-visual patterns in different frames of the dominant pedestrian vary with pose and viewpoint variations.

    Second, we further propose the Contrastive Feature Aggregation (CFA) module for robust temporal aggregation. In brief, CFA adaptively determines the contribution of each frame by exploring the temporal context from a contrastive perspective. More specifically, this module consists of two sequential steps. The first step estimates a frame-to-video similarity for each frame by aggregating its inter-frame relations with all frames in the video sequence. Afterwards, the second step determines the weight of each frame by simultaneously considering its own frame-to-video similarity and the average frame-to-video similarities of the other frames in the same video sequence. Therefore, compared with previous methods \cite{CVPR2017quality,ICCVChen2017joint,CVPR2020MGRAFA}, the frame weight in CFA is predicted in a contrastive manner, enabling the importance of comparatively better frames to be more effectively highlighted. Moreover, the proposed CFA bridges two types of feature aggregation methods, i.e. temporal pooling-based \cite{CVPR2017quality, CVPR2020MGRAFA} and information propagation-based \cite{CVPR2020MGH,CVPR2020STG,ICCVLi2019Global}. We show that the information propagation-based methods can be reduced to the temporal pooling-based ones, if information from the high-quality frames are forced to propagate to the poor-quality ones.

    Extensive experiments are conducted on four video-based ReID benchmarks, i.e. MARS \cite{ECCV2016mars}, DukeMTMC-VideoReID \cite{CVPR2018exploit,ristani2016MTMC}, iLIDS-VID \cite{ECCV2014vid}, and LS-VID \cite{ICCVLi2019Global}. Experimental results demonstrate the effectiveness of each component in CSA-Net and show that CSA-Net consistently achieves state-of-the-art performance on these databases.

\section{Related Work}
\label{sec:related}
    Early video-based ReID approaches tend to exploit temporal cues for video feature extraction. For example, McLaughlin et al. \cite{CVPRhe2016recurrent} utilized optical flow for video feature extraction. Li et al. \cite{AAAI2019multiscale} and Xu et al. \cite{ICCVChen2017joint} adopted 3D CNN and Recurrent Neural Network  respectively to extract spatial-temporal representations. However, these methods usually result in a large model size or high computational cost. Inspired by the success in image-based ReID~\cite{TIP2020cdpm,CVPRzheng2019re, CVPRhou2019interaction}, more recent works typically begin by extracting features from each individual frame, then adopt various strategies to aggregate the frame features into a single video feature \cite{ICCVLi2019Global,AAAI2019spatial, CVPR2020MGH,CVPR2020MGRAFA}. According to the way in which temporal aggregation is performed, existing approaches can be roughly divided into the following two categories.

    \textbf{Temporal Pooling-based Methods} An intuitive strategy involves applying temporal average pooling on frame features to obtain the video feature \cite{ECCV2016mars,CVPR2018exploit}. However, this strategy suffers from the impact of noisy frames. To address this issue, some works attempt to highlight features from high-quality frames and suppress those from low-quality ones by means of different weighting strategies \cite{CVPR2017quality,CVPR2017see,CVPR2018diversity, ICCVChen2017joint,CVPR2020MGRAFA,AAAI2018QEN}. For example, Liu et al. \cite{CVPR2017quality} introduced a quality aware network (QAN) to associate each frame with a quality score.
    Li et al. \cite{CVPR2018diversity} assigned weights to image regions, with each weight based on the visibility. Zhang et al. \cite{CVPR2020MGRAFA} predicted the weight for each frame feature according to its correlations with the averaged feature maps in a sequence. Although these approaches weaken the impact of noisy frames, they typically focus on the temporal aggregation step while underestimating the importance of the frame feature extraction step. Moreover, these approaches usually estimate the frame weight based on the content of each frame alone \cite{CVPR2017quality, CVPR2018diversity}, while tending to ignore the informative temporal cues. By contrast, we argue that the contribution of each frame should be determined with reference not only to its own quality, but also to the average quality of the other frames in the video sequence.

    \textbf{Information Propagation-based Methods} Methods in this category first refine each frame feature using the features of the other frames \cite{CVPR2020MGH,CVPR2020STG,ICCVLi2019Global,li2020spatial}. After the quality of each frame feature is improved, a simple temporal pooling strategy, such as averaging, can be adopted to obtain the video feature. For example, Li et al. \cite{ICCVLi2019Global} designed a temporal self-attention module to capture the long-term pair-wise relations between frames. The frame features are then revised according to the pair-wise relations between them. Besides, Yan et al. \cite{CVPR2020MGH} constructed multi-granular hypergraphs that model both the short- and long-term dependencies between frame features, thereby providing more diverse information to improve these frame features. However, when the self-attention mechanism is used, poor-quality frame features are more likely to be revised by semantically similar frames, the quality of which may also be poor. 

    Unlike the works discussed above, CSA-Net improves both the frame feature extraction and temporal aggregation steps; therefore, the extracted video feature is more discriminative and robust. Moreover, we propose a novel temporal aggregation method that not only achieves excellent performance, but also bridges the temporal pooling-based and information propagation-based methods.

\section{Context Sensing Attention Network}
\subsection{Overview}
\label{sec:overview}
     CSCA and CFA can be deployed on various baselines, including IDE \cite{zheng2017discriminatively}, PCB \cite{ECCVsun2018beyond}, and MPN \cite{wang2019mpn}. For the sake of simplicity, we illustrate the two modules on the IDE baseline in Figure \ref{framework}. In this IDE model, ResNet-50 \cite{CVPRhe2016deep} is adopted as the backbone; here, the last spatial down-sampling operation is removed, following \cite{ECCVsun2018beyond}, in order to increase the size of the output feature maps.

     As illustrated in Figure \ref{framework}, CSA-Net takes a sampled video sequence ${\bf{I}} = \{{\bf{I}}_{1}, {\bf{I}}_{2}, ... , {\bf{I}}_{T}\}$ as input. The backbone produces the frame-level feature maps $\{{\bf{F}}_{1}, {\bf{F}}_{2}, ... , {\bf{F}}_{T}\}$. CSCA then refines each ${\bf{F}}_{t}$, as follows:
     \begin{equation}
     \hat{\bf{F}}_{t} =  {\bf{F}}_{t} \otimes {\bf{c}}_{t}.
     \label{eq1}
     \end{equation}
     Here, $\otimes$ denotes the channel-wise multiplication operation. ${\bf{c}}_{t}$, which represents the channel weights obtained by CSCA for the $t$-th frame, considers both the individuality of the $t$-th frame and the overall content of the entire sequence.

     Next, the refined feature maps $\{\hat{\bf{F}}_{1}, \hat{\bf{F}}_{2}, ... , \hat{\bf{F}}_{T}\}$ are fed into a Global Average Pooling (GAP) layer and a Fully Connected (FC) layer to obtain the frame features $\{{\bf{f}}_{1}, {\bf{f}}_{2}, ... , {\bf{f}}_{T}\}$. Finally, these frame features are aggregated to form the video feature $\bf{h}$ via weighted averaging:
     \begin{equation}
     {\bf{h}} =  \frac{1}{T}\sum_{t=1}^T {{w}}_{t} {\bf{f}}_{t},
     \label{eq2}
     \end{equation}
     where ${{w}}_{t}$ denotes the weight estimated by the CFA module for the $t$-th frame. It is determined in a contrastive manner by considering both the quality of each individual frame and the average quality of the other frames in sequence.

     During training, both cross-entropy loss and triplet loss \cite{CVPR2015facenet} are employed to optimize $\bf{h}$, as illustrated in Figure~\ref{framework}. The two loss terms are realized in the same way as in existing works \cite{CVPR2020MGRAFA, CVPR2020MGH, CVPR2020STG}. During testing, $\bf{h}$ is employed as the representation of a video sequence. The cosine metric is adopted for performance evaluation.

\subsection{Context Sensing Channel Attention}

\begin{figure*}[t]
\centerline{\includegraphics[width=1.0\textwidth]{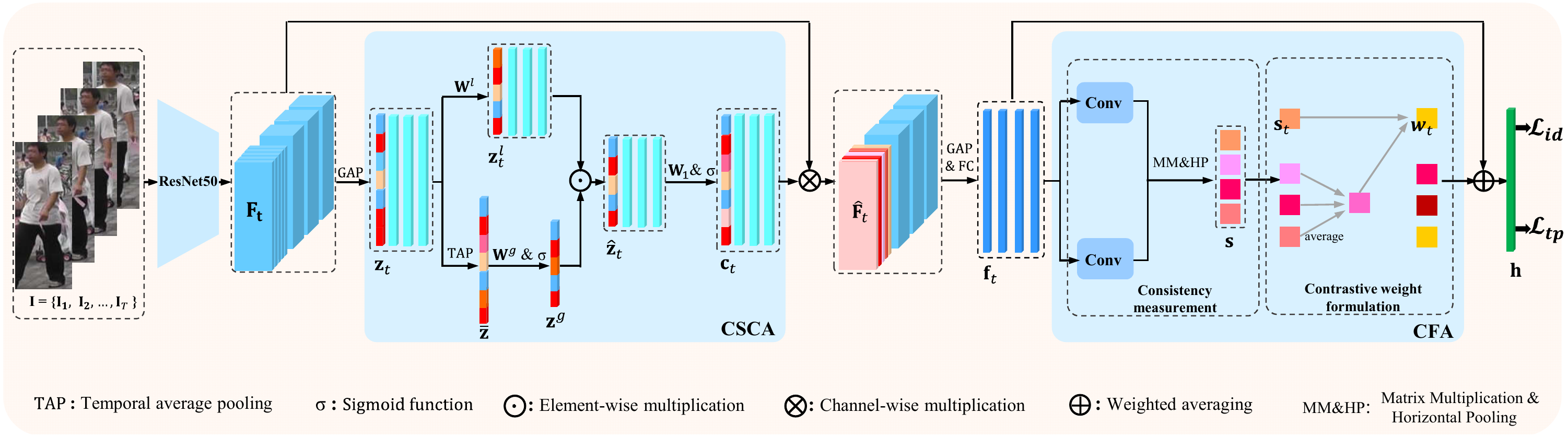}}
\caption{Architecture of CSA-Net. It includes two novel components, i.e. CSCA and CFA. CSCA is attached directly to the output feature maps of one backbone model, e.g. ResNet-50. It produces frame-level channel weights that highlight responses from relevant channels to the dominant pedestrian in the video sequence; in this way, it promotes the quality of frame features. CFA predicts the weights for frame features and aggregates these features into a single video feature via weighted averaging. The frame weight is computed in a contrastive manner, meaning that it is determined by both the quality of each individual frame and the average quality of the other frames in the sequence. Both cross-entropy loss and triplet loss are employed to optimize the video feature.}
\label{framework}
\end{figure*}

\label{sec:ca}
    We propose CSCA to highlight responses from informative channels for each frame and suppress those of less useful channels. As illustrated in Figure~\ref{problem}(a), image content may vary dramatically across frames, especially when interference exists between pedestrians, which makes it difficult to infer informative channels from each individual frame alone. 
    Fortunately, most frames in a video sequence are free from interference, which inspires us to infer informative channels for each image with the help of global temporal context in the video sequence.

    Accordingly, CSCA is designed with two key criteria in mind. First, informative channels for each frame should be relevant to the dominant pedestrian in the sequence. Second, informative channels can vary across frames in the same sequence; this is because of changes in pedestrian appearance due to pose and viewpoint variations. Therefore, CSCA aims to consider both the global context of the entire sequence and the individuality of each frame.

    The architecture of CSCA is illustrated in Figure~\ref{framework}. First, we feed the feature maps for each frame to a GAP layer, the obtained feature vector is denoted as ${\bf{z}}_{t}$ for the $t$-th frame. Second, CSCA adopts a two-branch structure: one for individual frames and the other for the entire sequence. The first branch processes ${\bf{z}}_{t}$ using a $1 \times 1$ Conv layer, the parameters of which are denoted as ${\bf{W}}^{l}$. The output of this layer is denoted as ${\bf{z}}_{t}^{l}$. The second branch comprises a temporal average pooling (TAP) layer, one $1 \times 1$ Conv layer, and one sigmoid layer. The two Conv layers do not share parameters. The operation of the second branch can be represented as follows:
    \begin{equation}
    \begin{aligned}
    \bar{\bf{z}} =& \frac{1}{T} \sum_{t=1}^T {\bf{z}}_{t}, \\
    {\bf{z}}^{g} = & \sigma ({\bf{W}}^{g} \bar{\bf{z}}),
    \label{ca2}
    \end{aligned}
    \end{equation}
    where $\sigma$ denotes the sigmoid function. ${\bf{W}}^{g} \in {\mathbb{R}}^{\frac{C}{r_{1}} \times C }$ denotes the parameters of the Conv layer in the second branch, while $r_{1}$ and $C$ denote the reduction ratio and the dimension of ${\bf{z}}_{t}$, respectively. Here, $r_{1}$ is empirically set to 4.

    Third, ${\bf{z}}^{g}$ acts as a gating mechanism to modulate elements in ${\bf{z}}_{t}^{l}$:
    \begin{equation}
    \hat{\bf{z}}_{t} = {\bf{z}}^{g} \odot {\bf{z}}_{t}^{l},
    \label{ca3}
    \end{equation}
    where $\odot$ represents the element-wise multiplication operation, which highlights the responses of relevant elements in ${\bf{z}}_{t}^{l}$ to the dominant pedestrian in the sequence and suppresses responses caused by interferences.

    Finally, we employ another $1 \times 1$ Conv layer, which is followed by a sigmoid layer, to obtain the final channel attention for the $t$-th frame:
    \begin{equation}
    {\bf{c}}_{t} = \sigma ({\bf{W}}_{1} \hat{\bf{z}}_{t}),
    \label{ca4}
    \end{equation}
    where ${\bf{W}}_{1} \in {\mathbb{R}}^{ C \times \frac{C}{r_{1}}}$ denotes the parameters of the Conv layer. ${\bf{c}}_{t}$ is used to refine ${\bf{F}}_{t}$ according to Equation (\ref{eq1}).

    Compared with the SE module~\cite{cvprse2018}, CSCA only introduces one extra $1 \times 1$ Conv layer; therefore, CSCA is still computationally efficient with a compact model structure.

    {\bf {\emph{Discussion}}} The channel weights obtained by CSCA for each frame are modulated by the content of the entire video sequence. Modulation occurs on ${\bf{z}}_{t}^{l}$ rather than ${\bf{c}}_{t}$. According to the difference in dimensionality, the elements in ${\bf{z}}_{t}^{l}$ and ${\bf{c}}_{t}$ can be interpreted as responses to macro- and micro-visual patterns, respectively. Intuitively, it is easier to infer the identity-relevant macro-visual patterns than the micro ones; this is because the latter may represent the individuality of one frame, while the former are usually stable across frames. In the experimentation section, we empirically prove that it is indeed better to impose the modulation on ${\bf{z}}_{t}^{l}$ rather than ${\bf{c}}_{t}$.

\subsection{Contrastive Feature Aggregation}
\label{sec:da}
    As illustrated in Figure \ref{problem}(c, d), the feature quality of some frames is inherently limited due to occlusion and pedestrian detection errors. We accordingly further propose the CFA module to weaken the influence of poor-quality frames by exploring temporal context information in the temporal aggregation step.

    Our key observation here is that, in most videos, only a small fraction of frames are of poor quality, the overall content in video sequence is therefore stable. This indicates that the consistency between each frame and all frames in sequence can reflect the quality of the individual frame. Inspired by this observation, we design the CFA module. In general, this module comprises two components, i.e. consistency measurement and contrastive weight formulation.

    \subsubsection{Consistency Measurement.} This component takes the frame features $\{{\bf{f}}_{1}, {\bf{f}}_{2}, ... , {\bf{f}}_{T}\}$ as input and estimates a quality score for each frame. This score, denoted as $s_{t}$ for the $t$-th frame, measures the average similarity between the $t$-th frame and each of the frames in sequence. As shown in Figure \ref{framework}, this component constructs two parallel $1 \times 1$ Conv layers that share parameters. The output of these two Conv layers is used to compute ${\bf{X}} \in \mathbb{R}^{{T} \times {T}}$ as follows:
    \begin{equation}
    x_{ij} = \frac{{{\theta}({\bf{{f}}}_i)}^{\mathsf{T}} {\theta}({\bf{{f}}}_j)} {\| {{\theta}({\bf{{f}}}_i)} \| \|{\theta}({\bf{{f}}}_j) \|},
    \label{matrix}
    \end{equation}
    where $x_{ij}$ denotes the element in the $i$-th row and $j$-th column of ${\bf{X}}$. It represents the cosine similarity between the two frame features ${\bf{{f}}}_i$ and ${\bf{{f}}}_j$. Moreover, ${{\theta}({\bf{{f}}}_{i})} = {\bf{W}}_{2} {\bf{{f}}}_{i}$, where ${\bf{W}}_{2} \in \mathbb{R}^{\frac{d}{r_{2}} \times d}$ represents the parameters of the Conv layer. $d$ denotes the dimension of frame features, while ${r_{2}}$ represents the reduction ratio, which is empirically set to 2.

    Finally, the consistency-aware quality score $s_{t}$ for the $t$-th frame can be obtained as follows:
    \begin{equation}
    s_{t} = \frac{1}{T} \sum_{i=1}^T x_{ti}.
    \label{similarity}
    \end{equation}

    \subsubsection{Contrastive Weight Formulation.} The next step is to aggregate the frame features. As introduced in Section \ref{sec:related}, recent information propagation-based methods \cite{CVPR2020MGH,CVPR2020STG,ICCVLi2019Global} first refine frame features according to their pair-wise relations, then apply a simple temporal averaging operation on the refined frame features to obtain the video feature. However, frames with close pair-wise relations tend to be of similar quality. Accordingly, we propose to adaptively improve each frame feature with reference to the high-quality ones as follows:
    \begin{equation}
    {\bf{\hat{f}}}_{t} = s_{t} {\bf{{f}}}_{t} +  (1 - s_{t}) \frac{1}{T-1} \sum_{\substack{i=1...T; i \neq t}} s_{i} {\bf{{f}}}_{i},
    \label{aggregation}
    \end{equation}
    where ${\bf{\hat{f}}}_{t}$ denotes the refined frame feature for the $t$-th frame. ${\bf{\hat{f}}}_{t}$ is made up of two parts. The first part, i.e. $s_{t} {\bf{{f}}}_{t}$, denotes the contribution of its original feature ${\bf{{f}}}_{t}$. Here, a larger value of $s_{t}$ indicates a greater contribution from ${\bf{{f}}}_{t}$. The second part introduces the contributions from the other $T$-1 frames according to their respective quality scores. Both parts work cooperatively to improve the original frame features.

    Finally, the video feature ${\bf{{h}}}$ is obtained by applying the temporal averaging on the improved frame features ${\bf{\hat{f}}}_{t}$:
    \begin{equation}
    \begin{aligned}
    {\bf{{h}}} & = \frac{1}{T} \sum_{t=1}^T \Big \{ s_{t} {\bf{{f}}}_{t} +  (1 - s_{t}) \frac{1}{T-1} \sum_{\substack{i=1...T; i \neq t}} s_{i} {\bf{{f}}}_{i} \Big \} \\
                       &= \frac{1}{T} \sum_{t=1}^T \Big \{ s_{t} \big( 2 - \frac{1}{T-1} \sum_{\substack{i=1...T; i \neq t}} s_{i} \big) \Big \} {\bf{{f}}}_{t}.
    \label{aggregation2}
    \end{aligned}
    \end{equation}

    Please refer to Equation (\ref{aggregation4}) for derivation of Equation (\ref{aggregation2}). Accordingly, the final weight for the $t$-th frame feature, i.e. $w_{t}$, can be decoupled into two parts: $s_{t}$ and $( 2 - \frac{1}{T-1} \sum_{i=1...T; i \neq t} s_{i})$. The first of these represents an absolute weight for ${\bf{{f}}}_{t}$. The second indicates the average quality score of all the other frames. This part enables $w_{t}$ to be determined in a contrastive manner: a higher value indicates that the average quality of the other frames is low, and thus that the importance of the $t$-th frame should be further emphasized. These two parts are complementary to each other and work collaboratively for temporal aggregation.

    {\bf {\emph{Discussion}}} Equation (\ref{aggregation}) and Equation (\ref{aggregation2}) bridge the two categories of temporal aggregation methods reviewed in Section \ref{sec:related}. They prove that the information propagation-based methods can be equivalent to the temporal pooling-based methods if poor-quality frame features are refined using high-quality ones. Moreover, unlike existing temporal pooling-based methods, CFA determines the frame weight in a contrastive manner; as demonstrated in the experimentation section, this is a more effective approach.

    {\bf {\emph{Derivation of Equation (\ref{aggregation2})}}} The video feature ${\bf{{h}}}$ is obtained as follows:
    \begin{equation}
        {\bf{{h}}} = \frac{1}{T} \sum_{t=1}^T \Big \{s_{t} {\bf{{f}}}_{t} +  (1 - s_{t}) \frac{1}{T-1} \sum_{\substack{i=1...T; i \neq t}} s_{i} {\bf{{f}}}_{i} \Big \}.
    \label{aggregation3}
    \end{equation}
    Accordingly, the weight $w_{t}$ for the $t$-th frame feature (\emph{i.e.} ${\bf{{f}}}_{t}$) can be formulated as follows:
    \begin{equation}
    \begin{aligned}
        w_{t} & =  s_{t} +  \sum_{\substack{i=1...T; i \neq t}} (1 - s_{i}) \frac{1}{T-1}  s_{t} \\
              & =  s_{t} +  \frac{1}{T-1}  s_{t} \sum_{\substack{i=1...T; i \neq t}} (1 - s_{i}) \\
              & =  s_{t} \Big \{ 1 +  \frac{1}{T-1} \sum_{\substack{i=1...T; i \neq t}} (1 - s_{i})  \Big \}\\
              & =  s_{t} \Big \{ 1 +  \frac{1}{T-1} \big \{ (T-1) - \sum_{\substack{i=1...T; i \neq t}} s_{i} \big \}  \Big \}\\
              & =  s_{t} ( 2 - \frac{1}{T-1} \sum_{\substack{i=1...T; i \neq t}} s_{i}  ).\\
    \label{aggregation4}
    \end{aligned}
    \end{equation}

\subsection{Ablation Study}
We systematically investigate the effectiveness of each key component of CSA-Net on LS-VID, DukeMTMC-VideoReID and MARS. To ensure comprehensive evaluation, both IDE and MPN~\cite{wang2019mpn} are adopted as baselines (for details of MPN, please refer to the appendix). Experimental results are summarized in Table~\ref{Ablation1}.

\newcommand{\tabincell}[2]{\begin{tabular}{@{}#1@{}}#2\end{tabular}}
\begin{table*}[t]
\caption{Ablation study on each key component of CSA-Net.}
\begin{tabular*}{\textwidth}{c|cccc|cccc|cccc}
  \thickhline
  \multirow{3}*{\tabincell{c}{Method}} & \multicolumn{4}{c|}{LS-VID} & \multicolumn{4}{c|}{DukeMTMC-VideoReID} & \multicolumn{4}{c}{MARS} \\\cline{2-13}
  & \multicolumn{2}{c}{IDE} & \multicolumn{2}{c|}{MPN} &\multicolumn{2}{c}{IDE} & \multicolumn{2}{c|}{MPN} &\multicolumn{2}{c}{IDE} & \multicolumn{2}{c}{MPN} \\
  \cline{2-13}
                          &Rank-1&mAP     &Rank-1&mAP           &Rank-1&mAP    &Rank-1&mAP      &Rank-1&mAP     &Rank-1&mAP     \\
  \hline
  \hline
  Baseline                      &77.5 &63.6 &82.8  &70.7        &96.0&94.7 &96.2 &95.4  &87.2&80.7  &88.5  &82.5\\

  CSCA                          &81.3 &68.1 &84.9  &72.8        &97.1&95.8 &97.4 &96.4  &88.7&82.7  &89.7  &84.1\\
  CFA                           &80.9 &67.4 &84.6  &72.5        &96.9&95.6 &97.3 &96.4  &88.4&82.6  &89.6  &83.9\\
  CSA-Net                       &82.5 &70.2 &85.3  &73.4        &97.4&96.2 &97.7 &96.7  &89.0&83.2  &90.4  &84.5\\
  \thickhline
\end{tabular*}
\label{Ablation1}
\end{table*}

\section{Experiments}
We evaluate our approach on four challenging benchmarks, namely MARS~\cite{ECCV2016mars}, DukeMTMC-VideoReID~\cite{CVPR2018exploit,ristani2016MTMC}, iLIDS-VID~\cite{ECCV2014vid}, and LS-VID~\cite{ICCVLi2019Global}, by following their respective evaluation protocols. The cumulative matching characteristic (CMC) and mean Average Precision (mAP) are adopted as evaluation metrics.

\textbf{Mars Dataset.} The Mars dataset is a large-scale benchmark for person ReID. It is an extension of the Market-1501 dataset \cite{zheng2015scalable} and comprises 17,503 video sequences belonging to 1,261 identities as well as 3,248 distractor sequences. Videos in this dataset were captured by 6 cameras. Pedestrians were detected using Deformable Part Models~\cite{felzenszwalb2009object}.

\textbf{DukeMTMC-VideoReID dataset.} includes 4,832 video sequences associated with 1,812 identities. Videos in this database were captured by 8 cameras. Bounding boxes of pedestrians were manually annotated.

\textbf{iLIDS-VID dataset.} The iLIDS-VID dataset consists of 600 video sequences of 300 identities. Two indoor cameras were utilized to capture the pedestrian sequences. Each video sequence contains 23 to 192 frames. This dataset is very challenging because of the large
variations on lighting and viewpoints and cluttered backgrounds.

\textbf{LS-VID dataset.} includes 14,943 video sequences of 3,772 identities. A camera network consisting of 3 outdoor cameras and 12 indoor cameras were employed  to construct this dataset.

\subsection{Implementation Details}
We implement the proposed CSA-Net based on the PyTorch framework. A standard stochastic gradient descent optimizer with a weight decay of $5 \times 10^{-4}$ and a momentum value~\cite{sutskever2013importance} of 0.9 is utilized for model optimization. Fine-tuned from the IDE model~\cite{CVPRhe2016deep}, CSA-Net is trained in an end-to-end fashion for 350 epochs on each of the four benchmarks, with the learning rate initially set to 0.01 and then multiplied by 0.1 every 100 epochs.

All images are resized to $256 \times 128$ pixels. For data augmentation, we only adopt random erasing~\cite{zhong2017random} with a ratio of 0.5. The margin of the triplet loss is empirically set to 0.25. We sample 4 video sequences for each of the 4 identities to construct a mini-batch; therefore, the batch size is 16. The values of $d$ and $C$ are 512 and 2048, respectively. During training, we set $T$ to 8; specifically, we sample 8 frames from each sequence by uniformly splitting the sequence into 4 segments and randomly sample 2 frames per segment. During testing, we use all frames of a video to generate the video feature if its length is less than 128; otherwise, 128 frames are sampled following the strategy discussed above.

\begin{figure*}[t]
    \centerline{\includegraphics[width=0.90\textwidth]{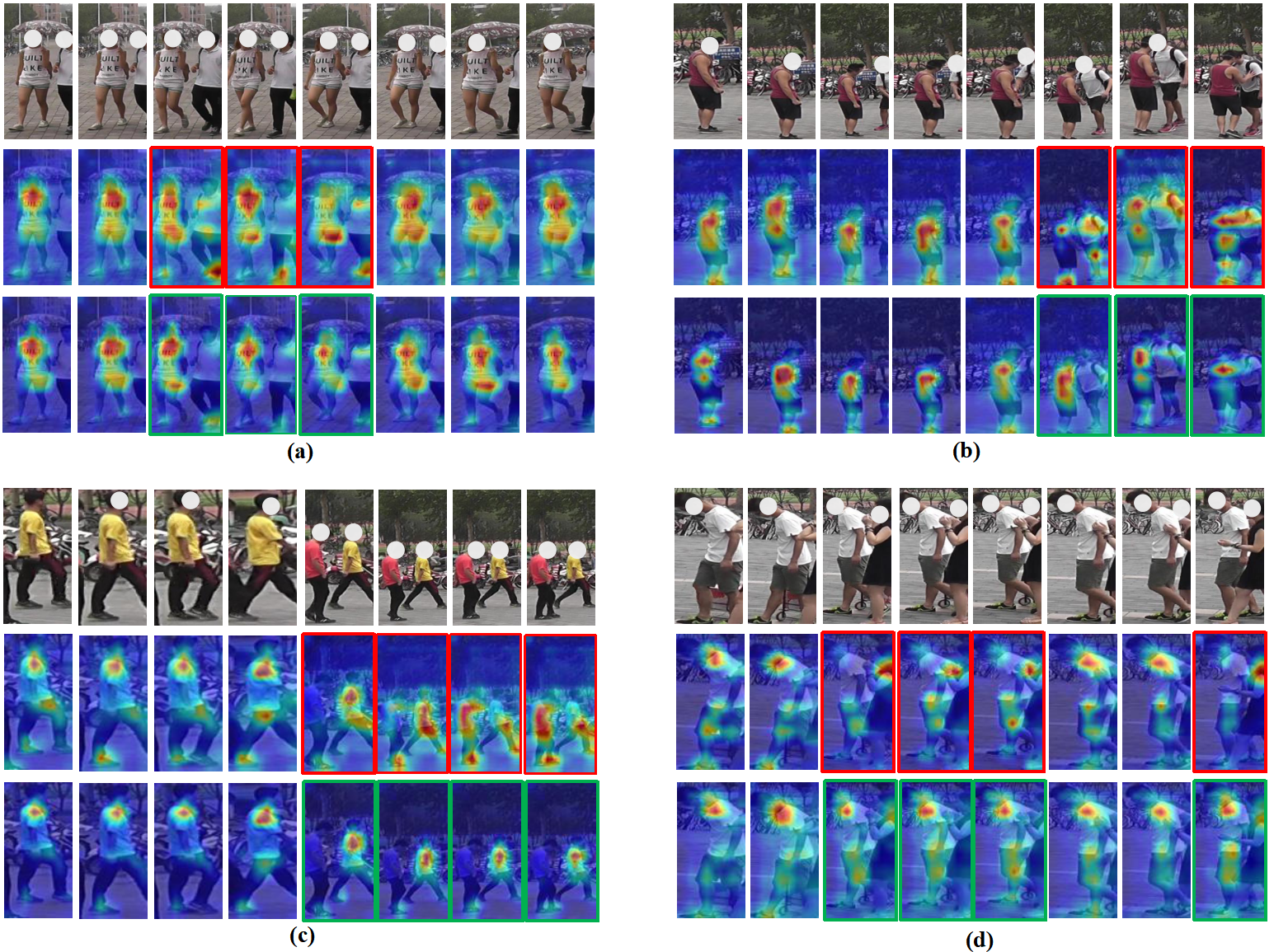}}
     \caption{Visualization of heat maps for feature maps produced by the IDE baseline (images in the second row) and CSCA-equipped IDE (images in the third row). Images in which the IDE baseline fails to focus on the dominate pedestrian are framed in red rectangles.}
    \label{grad}
\end{figure*}

\subsubsection{Effectiveness of CSCA}
In this experiment, we equip the baseline with CSCA only. As shown in Table~\ref{Ablation1}, CSCA yields clear performance improvements across all settings. For example, compared with the IDE baseline, CSCA improves the performance by 3.8\% and 4.5\% in terms of Rank-1 accuracy and mAP on LS-VID, respectively.

We further support the above quantitative results by visualizing the heat maps for frame-level feature maps produced by the IDE baseline and CSCA-equipped IDE, respectively. As illustrated in the second row of Figure~\ref{grad}, the feature maps produced by IDE show strong responses in the interference regions; by contrast, as the third row of Figure~\ref{grad} shows, CSCA robustly highlights the body region of the dominant pedestrian in the video sequence. These results demonstrate the effectiveness of CSCA.

\subsubsection{Effectiveness of CFA}
In this experiment, we equip the baseline with CFA only. The results listed in Table~\ref{Ablation1} show that CFA brings consistent performance promotion for both baselines. For example, compared with the IDE baseline, CFA improves the Rank-1 accuracy by 3.4\% and mAP by 3.8\% on LS-VID.

\begin{figure*}[t]
    \centerline{\includegraphics[width=0.75\textwidth]{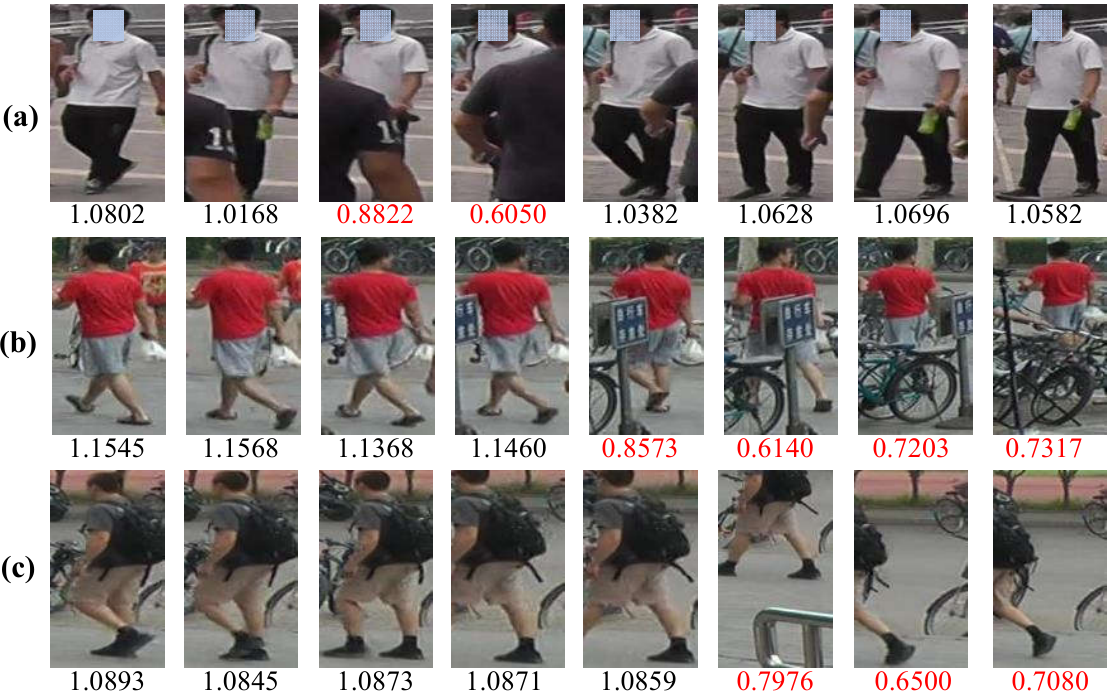}}
     \caption{Sampled video sequences with frame weights predicted by CFA. Each row stands for one sequence.}
    \label{weight}
\end{figure*}

To further support the above results, we examine the weights learned by CFA with the IDE baseline. As illustrated in Figure~\ref{weight}, the weights produced by CFA are reasonable; for example, CFA assigns lower weights to frames in which part missing or occlusions are present. These results justify the effectiveness of CFA.

Finally, we equip the baseline with both CSCA and CFA; this model is referred to as CSA-Net in Table~\ref{Ablation1}. We can observe from the table that CSA-Net consistently outperforms all other models in Table~\ref{Ablation1}. These comparisons validate that CSCA and CFA complement each other.

\subsection{Further Analysis and Discussions}
\subsubsection{CSCA Vs. SE}
We compare the performance of CSCA with the popular SE module~\cite{cvprse2018}. In the interest of efficient evaluation, only models based on the MPN baseline are evaluated. Two possible designs are compared, denoted as ``SE-frame" and ``SE-video" respectively in Table~\ref{Ablation2}. ``SE-frame" adopts an ordinary SE module to produce channel weights, i.e. ${\bf{c}}_{t}$, for each respective frame. Informative temporal cues of the sequence are ignored. ``SE-video" produces unified channel weights for all frames in the sequence. In more detail, the feature maps (i.e. ${\bf{F}_{t}}$) of all $T$ frames are first temporally averaged, after which channel weights for the averaged feature maps are extracted using an SE module. Subsequently, all ${\bf{F}_{t}}$ are refined using the obtained channel weights; therefore, ``SE-video" ignores the individuality of single frames. To facilitate fair comparison, the structure of the SE modules in both ``SE-frame" and ``SE-video" are the same as that for CSCA.

As shown in Table~\ref{Ablation2}, the performance of both ``SE-frame" and ``SE-video" is inferior to that of CSCA. For example, CSCA outperforms ``SE-frame" on LS-VID by 1.5\% and 1.3\% in terms of Rank-1 accuracy and mAP respectively. This is because CSCA not only considers the individuality of each frame, but also the overall content of the entire sequence; as a result, the channel weights produced by CSCA are more reasonable. These results demonstrate the superiority of CSCA.

\subsubsection{Comparisons with Variant of CSCA}
In Table~\ref{Ablation2}, we compare the performance of CSCA with one possible variant. This variant, denoted as ``CSCA-v", adopts two SE modules to generate frame- and video-level channel weights, respectively; the ways to learn frame- and video-level channel weights are the same as those adopted in ``SE-frame" and ``SE-video", respectively. The video-level channel weights are subsequently used to modulate each frame-level channel weights via element-wise multiplication. Accordingly, the essential difference between ``CSCA-v" and CSCA lies in the modulation position. To facilitate fair comparison, the other implementation details of ``CSCA-v" are kept the same as that in CSCA.

After assessing the results presented in Table~\ref{Ablation2}, we conclude that it is more effective to modulate channel weights in the hidden (first) Conv layer of the SE module. This may be because, compared with the output layer of the SE module, this hidden layer is more compact; its elements can thus be regarded to stand for macro-visual patterns, which are more coherent across frames in a video sequence. These comparisons demonstrate the effectiveness of CSCA.

\begin{figure*}[t]
    \centerline{\includegraphics[width=0.80\textwidth]{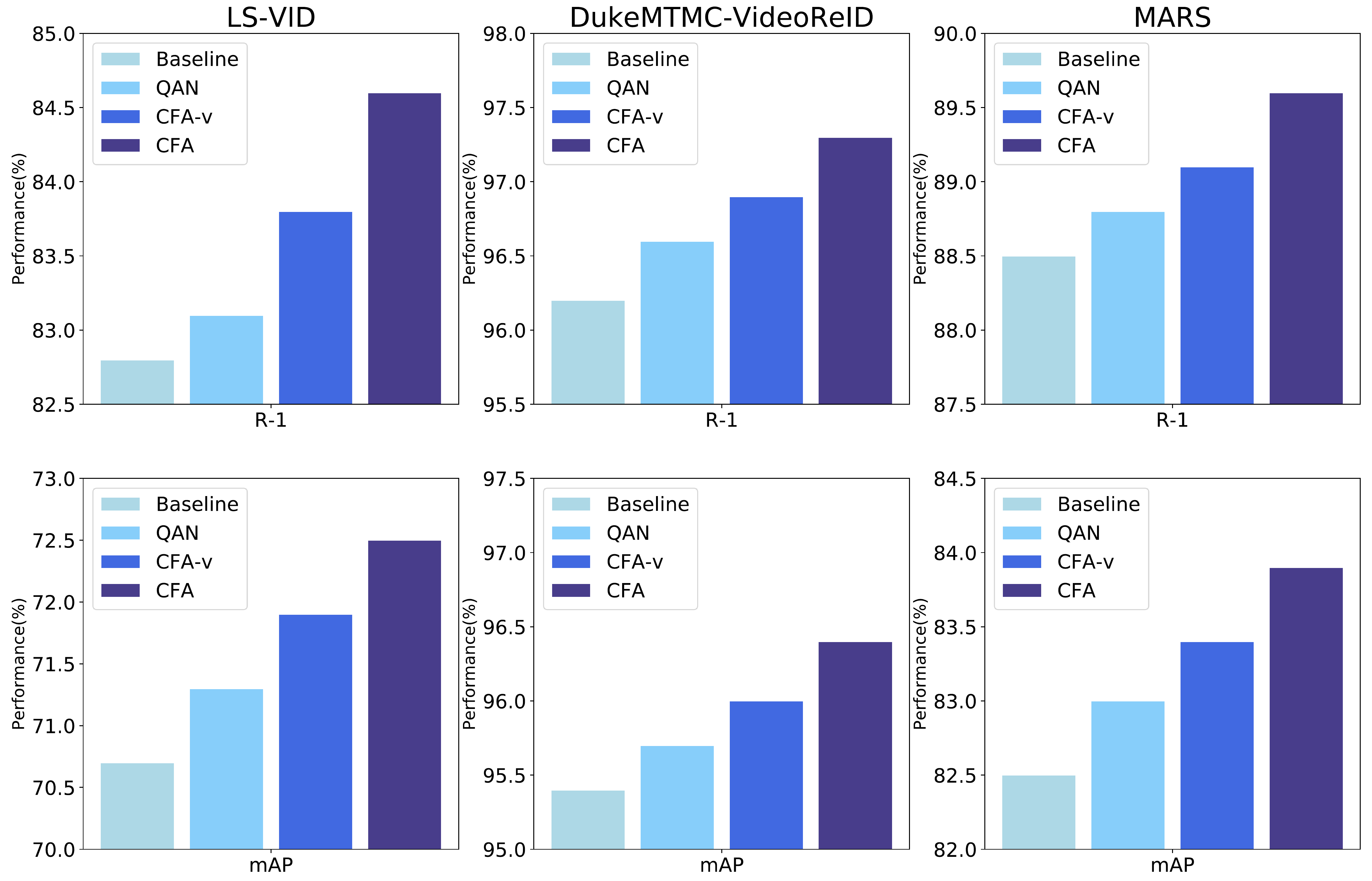}}
     \caption{Performance comparisons with QAN and one variant.}
    \label{da_v}
\end{figure*}

\subsubsection{Comparisons with QAN and Variant of CFA}
We compare the performance of CFA with QAN and one possible variant. The variant is denoted as ``CFA-v" in Figure~\ref{da_v}. QAN computes the weight for each frame based on its own content using an approach adopted in~\cite{CVPR2017quality}. In brief, it comprises a $1 \times 1$ Conv layer with an output dimension of 1 and a sigmoid layer for normalization. The input to QAN is the individual feature of each frame. For its part, ``CFA-v" adopts the frame-to-video similarity $s_{t}$ computed in Equation (\ref{similarity}) as the weight for the $t$-th frame.

From the comparisons presented in Figure~\ref{da_v}, we can make the following observations. First, both QAN and ``CFA-v" outperform the baseline, which demonstrates the effectiveness of weighting strategy. Second, ``CFA-v" beats QAN. This is because the weight predicted by ``CFA-v" is based on temporal cues, meaning that it is easier for ``CFA-v" to identify frames that have been contaminated by interference. Third, CFA surpasses ``CFA-v". This result validates the superiority of the contrastive weighting strategy, which considers both the quality of an individual frame and the average quality of the other frames in sequence.

\begin{figure*}[t]
    \centerline{\includegraphics[width=0.95\textwidth]{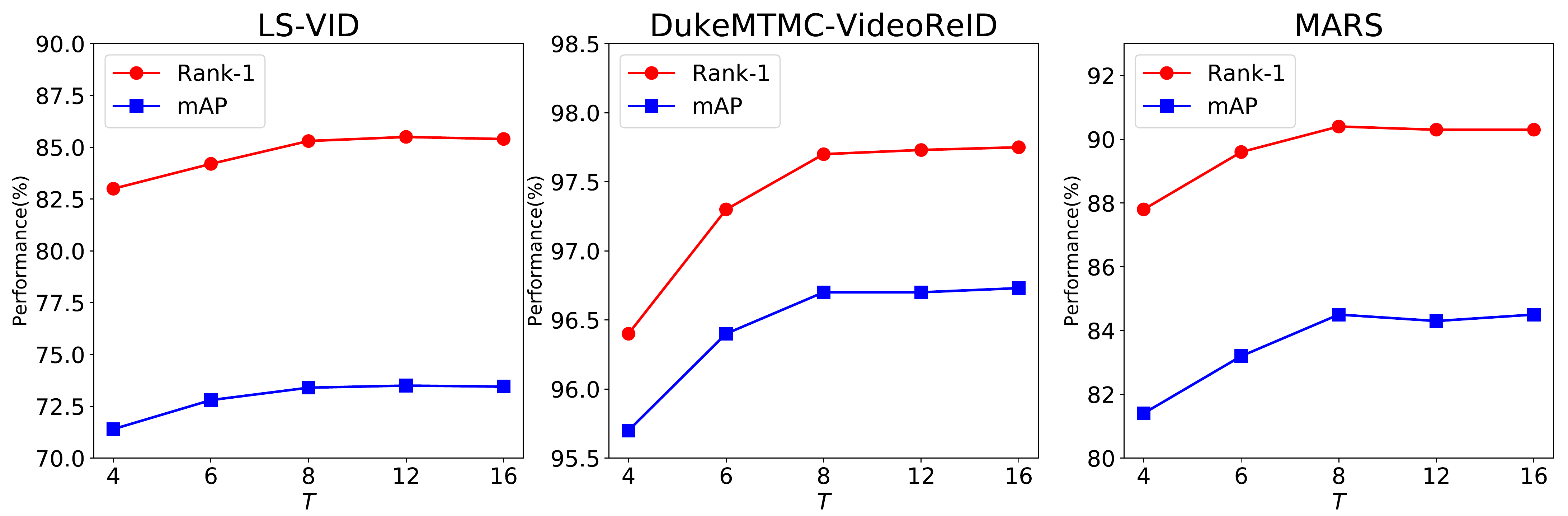}}
     \caption{Evaluation of the value of hyper-parameter $T$.}
    \label{T}
\end{figure*}

\subsubsection{CFA Vs. Non-local}
As illustrated in Equation (\ref{aggregation}), the proposed CFA module also works in the manner of information propagation. Therefore, in this experiment, we compare CFA with one of the most popular information propagation-based method, i.e. the non-local module~\cite{CVPRwang2017nonlocal}. Implementation of the non-local module follows~\cite{ICCVLi2019Global}. The other experimental settings remain unchanged to facilitate a clean comparison.

\begin{table*}[t]
\caption{Performance comparisons with the SE module and one variant of CSCA.}
\centering
\begin{center}
\begin{tabular}{c|cccccc}
  \thickhline
  \multirow{2}*{\tabincell{c}{Method}} & \multicolumn{2}{c}{LS-VID} & \multicolumn{2}{c}{Duke-Video} &\multicolumn{2}{c}{MARS}\\ \cline{2-7}

                                 &Rank-1&mAP          &Rank-1&mAP       &Rank-1&mAP\\
  \hline
  \hline
  Baseline                       &82.8  &70.7   &96.2 &95.4   &88.5  &82.5 \\
  \hline
  CSCA-v                         &84.2  &72.3   &97.0 &96.1   &89.3  &83.5 \\
  SE-frame                       &83.4  &71.5    &96.6 &95.7   &88.7  &82.6  \\
  SE-video                       &83.9  &71.9    &96.7 &95.9   &89.0  &83.0  \\
  \hline
  CSCA                           &84.9  &72.8    &97.4 &96.4   &89.7  &84.1  \\
  \thickhline
\end{tabular}
\end{center}
\label{Ablation2}
\end{table*}

\begin{table*}[t]
\caption{Performance comparisons with the non-local module.}
\centering
\begin{center}
\begin{tabular}{c|cccccc}
  \thickhline
  \multirow{2}*{\tabincell{c}{Method}} & \multicolumn{2}{c}{LS-VID} & \multicolumn{2}{c}{Duke-Video} &\multicolumn{2}{c}{MARS}\\ \cline{2-7}

                                 &Rank-1&mAP          &Rank-1&mAP       &Rank-1&mAP\\
  \hline
  \hline
  Baseline                      &82.8  &70.7      &96.2 &95.4  &88.5  &82.5   \\
  \hline
  Non-local                     &83.5  &71.3      &96.6 &95.7  &88.9  &82.8  \\
  CFA                           &84.6  &72.5      &97.3 &96.4   &89.6  &83.9  \\
  \thickhline
\end{tabular}
\end{center}
\label{nonlocal}
\end{table*}

From the results presented in Table~\ref{nonlocal}, we can observe that CFA consistently outperforms the non-local module on all three benchmarks. This is because the non-local module inherently tends to refine a poor-quality frame feature using these semantically similar frames, indicating that poor-quality frame features may be revised by ones of similar quality. In comparison, CFA adaptively improves poor-quality frame features with reference to the high-quality ones. These experimental results justify the superiority of CFA.

\subsubsection{Impact of the hyper-parameter T.}
In this experiment, we evaluate the performance of CSA-Net at different values of $T$ (namely 4, 6, 8, 12, and 16). All the other experimental settings remain unchanged to facilitate a clean comparison.

From the experimental results illustrated in Figure~\ref{T}, we can make the following observations. First, the performance of CSA-Net tends to be better at higher values of $T$; this is because a longer sequence provides more stable temporal context. Second, the performance of CSA-Net is robust to the value of $T$ if the value is sufficiently large. In light of these results, we set $T$ as 8 after considering the trade-off between performance and computational efficiency.

\subsection{Comparisons with State-of-the-Art Methods}

\begin{table*}[!t]
\caption{Performance comparisons on MARS~\cite{ECCV2016mars}, DukeMTMC-VideoReID~\cite{CVPR2018exploit,ristani2016MTMC}, and iLIDS-VID~\cite{ECCV2014vid}.}
\begin{tabular*}{\textwidth}{c|cccc|ccc|ccc}
\thickhline
  \multirow{2}*{\tabincell{c}{Method}} &\multicolumn{4}{c|}{MARS}  &\multicolumn{3}{c|}{Duke-Video}  &\multicolumn{3}{c}{iLIDS-VID} \\ \cline{2-11}
   &Rank-1 &Rank-5 &Rank-20 &mAP &Rank-1 &Rank-5 &mAP  &Rank-1 &Rank-5 &Rank-20\\
  \hline
  \hline
  Mars \cite{ECCV2016mars}             &68.3 &82.6 &89.4 &49.3  &-&-&-                            &53.0 &81.4  &95.1\\
  SeeForest \cite{CVPR2017see}         &70.6 &90.0 &-    &50.7  &-&-&-                            &55.2 &86.5  &97.0\\
  QAN \cite{CVPR2017quality}           &-&-&-&-                 &-&-&-                            &68.0 &86.8  &97.4\\
  RQEN \cite{AAAI2018QEN}              &77.8 &88.8 &94.3 &71.1  &-&-&-                            &80.0 &94.4  &99.3\\
  EUG \cite{CVPR2018exploit}           &80.8 &92.1 &96.1 &67.4  &83.6&94.6&78.3                   &-    &-     &-   \\
  CSA \cite{WACV2020Matiyali}          &83.4 &93.4 &97.4 &83.3  &89.3&98.3&88.5               &86.3 &97.4  &99.7\\
  TKP~\cite{ICCVGu2019Temporal}        &84.0 &93.7 &95.7 &73.3  &94.0&-   &91.7                   &-    &-      &-   \\
  M3D \cite{AAAI2019multiscale}        &84.4 &93.8 &97.7 &74.1  &-&-&-                            &74.0 &94.3   &-   \\
  COSAM~\cite{ICCVSubramaniam2019Co}   &84.9 &95.5 &97.9 &79.9  &95.4&99.3&94.1                   &79.6 &95.3   &-   \\
  Snippt~\cite{CVPR2018video}          &86.3 &94.7 &98.2 &76.1  &-&-&-                            &85.4 &96.7   &99.5\\
  STA~\cite{AAAI2019sta}               &86.3 &95.7 &-    &80.8  &96.2&99.3&94.9                   &-    &-      &    \\
  GLTR~\cite{ICCVLi2019Global}         &87.0 &95.8 &98.2 &78.5  &96.3&99.3&93.7                   &86.0 &98.0   &-   \\
  Attribute~\cite{CVPR2019attribute}   &87.0 &95.4 &98.7 &78.2  &-&-&-                            &86.3 &97.4   &99.7\\
  FGRA~\cite{AAAI2020frame}            &87.3 &96.0 &98.1 &81.2  &-&-&-                            &88.0 &96.7   &99.3\\
  VRSTC~\cite{CVPR2019vrstc}           &88.5 &96.5 &-    &82.3  &95.0&99.1&93.5                   &83.4 &95.5   &99.5\\
  MG-RAFA~\cite{CVPR2020MGRAFA}        &88.8 &97.0 &98.5 &85.9  &-&-&-                            &88.6 &98.0   &99.7\\
  TACAN \cite{WACV2020li}              &89.1 &96.1 &98.0 &84.0  &96.2 &99.4  &95.4    &88.9 &-      &-   \\
  TCLNet~\cite{ECCV2020Temporal}       &89.8 &-    &-    &85.1  &96.9&-&96.2                      &86.6 &-      &-   \\
  STGCN~\cite{CVPR2020STG}             &90.0 &96.4 &98.3 &83.7  &97.3&99.3  &95.7                &-    &-      &-   \\
  MGH~\cite{CVPR2020MGH}               &90.0 &96.7 &98.5 &85.8  &-&-&-                            &85.6 &97.1   &-   \\
  AP3D~\cite{ECCV2020Appearance}       &90.1 &-    &-    &85.1  &96.3&-&95.6                      &86.7 &-      &-   \\
  AFA~\cite{ECCV2020Coherence}         &90.2 &96.6 &-    &82.9  &97.2&99.4  &95.4     &88.5 &96.8   &-   \\
  BiCnet-TKS~\cite{CVPR2021TKS}        &90.2 &-    &-    &86.0  &96.3&-     &96.1     &-    &-      &-   \\
  STRF~\cite{aich2021spatio}           &90.3 &-    &-    &86.1  &97.4&-     &96.4     &89.3    &-      &-   \\

  \bfseries CSA-Net &{\bfseries 90.4} &96.7 & 98.5 & 84.5 &{\bfseries 97.7} &{\bfseries 99.4} &{\bfseries 96.7} &{\bfseries 90.0} &{\bfseries 98.3} &{\bfseries 99.8}\\
\thickhline
\end{tabular*}
\label{MARS}
\end{table*}

State-of-the-art approaches to video-based ReID usually extract part features~\cite{CVPR2020STG,CVPR2020MGH} or utilize backbones more powerful than ours (AP3D~\cite{ECCV2020Appearance} adopts a 3D-based ResNet-50 as the backbone, while the plain ResNet-50 is used in CSA-Net). To facilitate fair comparison, we adopt MPN~\cite{wang2019mpn} that extracts part features as the baseline in this subsection, since its performance is comparable to that obtained by baselines of recent methods~\cite{ECCV2020Coherence,ECCV2020Appearance}. Comparisons between CSA-Net and state-of-the-art methods are presented in Table~\ref{MARS}. From the table, it can be seen that CSA-Net consistently achieves state-of-the-art performance on each dataset. Specifically, on the DukeMTMC-VideoReID dataset, CSA-Net outperforms one of the most recent methods (i.e., AFA~\cite{ECCV2020Coherence}) by 0.5\% and 1.3\% in terms of Rank-1 accuracy and mAP, respectively. On the MARS database, CSA-Net achieves the best Rank-1 accuracy of 90.4\%. Besides, CSA-Net also surpasses state-of-the-art approaches on iLIDS-VID by at least 1.4\% in terms of Rank-1 accuracy. These comparisons demonstrate the superiority of CSA-Net.

Moreover, CSA-Net suppresses two recent part-based methods, i.e., STGCN~\cite{CVPR2020STG} and MGH~\cite{CVPR2020MGH}, on three datasets for the Rank-1 accuracy. Besides, compared with AP3D~\cite{ECCV2020Appearance}, which adopts 3D convolutions, the model structure of CSA-Net is simpler as all its operations are 2D-based.

Finally, we compare CSA-Net with recent approaches on the LS-VID~\cite{ICCVLi2019Global} dataset. As LS-VID was released only recently, few works have reported their performance on this dataset. It can be seen from Table~\ref{VID} that CSA-Net outperforms all comparison methods by significant margins in terms of Rank-1 accuracy. For example, CSA-Net outperforms BiCnet-TKS~\cite{CVPR2021TKS}, the most recent method, by 0.7\%. These experimental results are consistent with those on the first three datasets. In summary, the above comparisons further validate the effectiveness of CSA-Net.


\begin{table}[t]
\center
\caption{Performance comparisons on the LS-VID dataset~\cite{ICCVLi2019Global}.}
\begin{tabular}{p{2.0cm}<{\centering}|p{1.0cm}<{\centering}p{1.0cm}<{\centering}p{1.0cm}<{\centering}p{1.13cm}<{\centering}}
\thickhline
  {Methods} &R-1 &R-5 &R-20 &mAP \\ \cline{2-5}
\hline
\hline
  STMP~\cite{AAAI2019sta, ICCVLi2019Global}                   &56.8 &76.2  &87.1 &39.1\\
  M3D~\cite{AAAI2019multiscale, ICCVLi2019Global}         &57.7 &76.1  &88.2 &40.1\\
  PCB~\cite{ECCVsun2018beyond}             &75.7 &89.9 &94.7 &62.6\\
  GLTR~\cite{ICCVLi2019Global}                 &63.1 &77.2 &88.4 &44.3\\
  TCLNet~\cite{ECCV2020Temporal}                             &81.5 &- &- &70.3\\
  AFA~\cite{ECCV2020Coherence}                             &84.5 &- &- &73.2\\
  BiCnet-TKS~\cite{CVPR2021TKS}                            &84.6 &-     &-     &75.1\\
  \bfseries CSA-Net &{\bfseries 85.3} &{\bfseries 93.8} &{\bfseries 97.5} & 73.4\\
\thickhline
\end{tabular}
\label{VID}
\end{table}

%

\section{Conclusion}
In this paper, we propose a novel model, named CSA-Net, which improves both the frame feature extraction and temporal aggregation steps for robust video-based ReID. CSA-Net incorporates two novel components, i.e. Context Sensing Channel Attention (CSCA) and Contrastive Feature Aggregation (CFA). CSCA effectively highlights informative channels for each frame with reference to the content of the entire sequence. For its part, CFA predicts the weight of each frame for temporal aggregation; here, the weight is based on the coherence degree between each frame and the entire sequence, and is adaptively determined in a contrastive manner. Experimental results on four benchmarks demonstrate the effectiveness of CSA-Net.

{\small
\bibliographystyle{ieee_fullname}
\bibliography{sample-base}

\begin{thebibliography}{10}\itemsep=-1pt

\bibitem{aich2021spatio}
Abhishek Aich, Meng Zheng, Srikrishna Karanam, Terrence Chen, Amit~K
  Roy-Chowdhury, and Ziyan Wu.
\newblock Spatio-temporal representation factorization for video-based person
  re-identification.
\newblock In {\em Proceedings of the IEEE/CVF International Conference on
  Computer Vision}, pages 152--162, 2021.

\bibitem{CVPR2018video}
Dapeng Chen, Hongsheng Li, Tong Xiao, Shuai Yi, and Xiaogang Wang.
\newblock Video person re-identification with competitive snippet-similarity
  aggregation and co-attentive snippet embedding.
\newblock In {\em Proceedings of the IEEE/CVF Conference on Computer Vision and
  Pattern Recognition}, pages 1169--1178, 2018.

\bibitem{ECCV2020Coherence}
Guangyi Chen, Yongming Rao, Jiwen Lu, and Jie Zhou.
\newblock Temporal coherence or temporal motion: Which is more critical for
  video-based person re-identification?
\newblock In {\em European Conference on Computer Vision}, pages 660--676.
  Springer, 2020.

\bibitem{AAAI2020frame}
Zengqun Chen, Zhiheng Zhou, Junchu Huang, Pengyu Zhang, and Bo Li.
\newblock Frame-guided region-aligned representation for video person
  re-identification.
\newblock In {\em Proceedings of the AAAI Conference on Artificial
  Intelligence}, volume~34, pages 10591--10598, 2020.

\bibitem{wang2019mpn}
Changxing Ding, Kan Wang, Pengfei Wang, and Dacheng Tao.
\newblock Multi-task learning with coarse priors for robust part-aware person
  re-identification.
\newblock {\em IEEE Transactions on Pattern Analysis and Machine Intelligence},
  44(3):1474--1488, 2022.

\bibitem{fan2018unsupervised}
Hehe Fan, Liang Zheng, Chenggang Yan, and Yi Yang.
\newblock Unsupervised person re-identification: Clustering and fine-tuning.
\newblock {\em ACM Trans. Multimedia Comput. Commun. Appl.}, 14(4):1--18, 2018.

\bibitem{fang2021set}
Pengfei Fang, Pan Ji, Lars Petersson, and Mehrtash Harandi.
\newblock Set augmented triplet loss for video person re-identification.
\newblock In {\em Proceedings of the IEEE/CVF Winter Conference on Applications
  of Computer Vision}, pages 464--473, 2021.

\bibitem{felzenszwalb2009object}
Pedro~F Felzenszwalb, Ross~B Girshick, David McAllester, and Deva Ramanan.
\newblock Object detection with discriminatively trained part-based models.
\newblock {\em IEEE Transactions on Pattern Analysis and Machine Intelligence},
  32(9):1627 -- 1645, 2009.

\bibitem{AAAI2019sta}
Yang Fu, Xiaoyang Wang, Yunchao Wei, and Thomas Huang.
\newblock Sta: Spatial-temporal attention for large-scale video-based person
  re-identification.
\newblock In {\em Proceedings of the AAAI conference on artificial
  intelligence}, volume~33, pages 8287--8294, 2019.

\bibitem{gao2021mso}
Yajun Gao, Tengfei Liang, Yi Jin, Xiaoyan Gu, Wu Liu, Yidong Li, and Congyan
  Lang.
\newblock Mso: Multi-feature space joint optimization network for rgb-infrared
  person re-identification.
\newblock In {\em Proceedings of the 29th ACM International Conference on
  Multimedia}, pages 5257--5265, 2021.

\bibitem{ge2021cross}
Wenhang Ge, Chunyan Pan, Ancong Wu, Hongwei Zheng, and Wei-Shi Zheng.
\newblock Cross-camera feature prediction for intra-camera supervised person
  re-identification across distant scenes.
\newblock In {\em Proceedings of the 29th ACM International Conference on
  Multimedia}, pages 3644--3653, 2021.

\bibitem{ECCV2020Appearance}
Xinqian Gu, Hong Chang, Bingpeng Ma, Hongkai Zhang, and Xilin Chen.
\newblock Appearance-preserving 3d convolution for video-based person
  re-identification.
\newblock In {\em European Conference on Computer Vision}, pages 228--243.
  Springer, 2020.

\bibitem{ICCVGu2019Temporal}
Xinqian Gu, Bingpeng Ma, Hong Chang, Shiguang Shan, and Xilin Chen.
\newblock Temporal knowledge propagation for image-to-video person
  re-identification.
\newblock In {\em Proceedings of the IEEE/CVF International Conference on
  Computer Vision}, pages 9647--9656, 2019.

\bibitem{CVPRhe2016deep}
Kaiming He, Xiangyu Zhang, Shaoqing Ren, and Jian Sun.
\newblock Deep residual learning for image recognition.
\newblock In {\em Proceedings of the IEEE/CVF Conference on Computer Vision and
  Pattern Recognition}, pages 770--778, 2016.

\bibitem{CVPR2021TKS}
Ruibing Hou, Hong Chang, Bingpeng Ma, Rui Huang, and Shiguang Shan.
\newblock Bicnet-tks: Learning efficient spatial-temporal representation for
  video person re-identification.
\newblock In {\em Proceedings of the IEEE/CVF Conference on Computer Vision and
  Pattern Recognition}, pages 2014--2023, 2021.

\bibitem{ECCV2020Temporal}
Ruibing Hou, Hong Chang, Bingpeng Ma, Shiguang Shan, and Xilin Chen.
\newblock Temporal complementary learning for video person re-identification.
\newblock In {\em European Conference on Computer Vision}, pages 388--405.
  Springer, 2020.

\bibitem{CVPRhou2019interaction}
Ruibing Hou, Bingpeng Ma, Hong Chang, Xinqian Gu, Shiguang Shan, and Xilin
  Chen.
\newblock Interaction-and-aggregation network for person re-identification.
\newblock In {\em Proceedings of the IEEE/CVF Conference on Computer Vision and
  Pattern Recognition}, pages 9317--9326, 2019.

\bibitem{CVPR2019vrstc}
Ruibing Hou, Bingpeng Ma, Hong Chang, Xinqian Gu, Shiguang Shan, and Xilin
  Chen.
\newblock Vrstc: Occlusion-free video person re-identification.
\newblock In {\em Proceedings of the IEEE/CVF Conference on Computer Vision and
  Pattern Recognition}, pages 7183--7192, 2019.

\bibitem{cvprse2018}
Jie Hu, Li Shen, and Gang Sun.
\newblock Squeeze-and-excitation networks.
\newblock In {\em Proceedings of the IEEE/CVF Conference on Computer Vision and
  Pattern Recognition}, pages 7132--7141, 2018.

\bibitem{ICCVLi2019Global}
Jianing Li, Jingdong Wang, Qi Tian, Wen Gao, and Shiliang Zhang.
\newblock Global-local temporal representations for video person
  re-identification.
\newblock In {\em Proceedings of the IEEE/CVF International Conference on
  Computer Vision}, pages 3958--3967, 2019.

\bibitem{AAAI2019multiscale}
Jianing Li, Shiliang Zhang, and Tiejun Huang.
\newblock Multi-scale 3d convolution network for video based person
  re-identification.
\newblock In {\em Proceedings of the AAAI Conference on Artificial
  Intelligence}, volume~33, pages 8618--8625, 2019.

\bibitem{TIP2020limulti}
Jianing Li, Shiliang Zhang, and Tiejun Huang.
\newblock Multi-scale temporal cues learning for video person
  re-identification.
\newblock {\em IEEE Transactions on Image Processing}, 29:4461--4473, 2020.

\bibitem{WACV2020li}
Mengliu Li, Han Xu, Jinjun Wang, Wenpeng Li, and Yongli Sun.
\newblock Temporal aggregation with clip-level attention for video-based person
  re-identification.
\newblock In {\em Proceedings of the IEEE/CVF Winter Conference on Applications
  of Computer Vision}, 2020.

\bibitem{CVPR2018diversity}
Shuang Li, Slawomir Bak, Peter Carr, and Xiaogang Wang.
\newblock Diversity regularized spatiotemporal attention for video-based person
  re-identification.
\newblock In {\em Proceedings of the IEEE/CVF Conference on Computer Vision and
  Pattern Recognition}, pages 369--378, 2018.

\bibitem{li2020spatial}
Zhaoju Li, Zongwei Zhou, Nan Jiang, Zhenjun Han, Junliang Xing, and Jianbin
  Jiao.
\newblock Spatial preserved graph convolution networks for person
  re-identification.
\newblock {\em ACM Trans. Multimedia Comput. Commun. Appl.}, 16(1s):1--14,
  2020.

\bibitem{Trans2018pedestrian}
Hao Liu, Zequn Jie, Karlekar Jayashree, Meibin Qi, Jianguo Jiang, Shuicheng
  Yan, and Jiashi Feng.
\newblock Video-based person re-identification with accumulative motion
  context.
\newblock {\em IEEE Transactions on Circuits and Systems for Video Technology},
  28(10):2788¨C--2802, 2018.

\bibitem{liu2019dense}
Jiawei Liu, Zheng-Jun Zha, Xuejin Chen, Zilei Wang, and Yongdong Zhang.
\newblock Dense 3d-convolutional neural network for person re-identification in
  videos.
\newblock {\em ACM Trans. Multimedia Comput. Commun. Appl.}, 15(1s):1--19,
  2019.

\bibitem{liu2021viewing}
Liangchen Liu, Xi Yang, Nannan Wang, and Xinbo Gao.
\newblock Viewing from frequency domain: A dct-based information enhancement
  network for video person re-identification.
\newblock In {\em Proceedings of the 29th ACM International Conference on
  Multimedia}, pages 227--235, 2021.

\bibitem{liu2021watching}
Xuehu Liu, Pingping Zhang, Chenyang Yu, Huchuan Lu, and Xiaoyun Yang.
\newblock Watching you: Global-guided reciprocal learning for video-based
  person re-identification.
\newblock In {\em Proceedings of the IEEE/CVF Conference on Computer Vision and
  Pattern Recognition}, pages 13334--13343, 2021.

\bibitem{CVPR2017quality}
Yu Liu, Junjie Yan, and Wanli Ouyang.
\newblock Quality aware network for set to set recognition.
\newblock In {\em Proceedings of the IEEE/CVF Conference on Computer Vision and
  Pattern Recognition}, pages 5790--5799, 2017.

\bibitem{AAAI2019spatial}
Yiheng Liu, Zhenxun Yuan, Wengang Zhou, and Houqiang Li.
\newblock Spatial and temporal mutual promotion for video-based person
  re-identification.
\newblock In {\em Proceedings of the AAAI Conference on Artificial
  Intelligence}, volume~33, pages 8786--8793, 2019.

\bibitem{WACV2020Matiyali}
Neeraj Matiyali and Gaurav Sharma.
\newblock Video person re-identification using learned clip similarity
  aggregation.
\newblock In {\em Proceedings of the IEEE/CVF Winter Conference on Applications
  of Computer Vision}, pages 2655--2664, 2020.

\bibitem{CVPRhe2016recurrent}
Niall McLaughlin, Jesus~Martinez Del~Rincon, and Paul Miller.
\newblock Recurrent convolutional network for video-based person
  re-identification.
\newblock In {\em Proceedings of the IEEE/CVF Conference on Computer Vision and
  Pattern Recognition}, pages 1325--1334, 2016.

\bibitem{pang2022fully}
Bo Pang, Deming Zhai, Junjun Jiang, and Xianming Liu.
\newblock Fully unsupervised person re-identification via selective contrastive
  learning.
\newblock {\em ACM Trans. Multimedia Comput. Commun. Appl.}, 18(2):1--15, 2022.

\bibitem{ECCV2020Exploiting}
Dripta~S Raychaudhuri and Amit~K Roy-Chowdhury.
\newblock Exploiting temporal coherence for self-supervised one-shot video
  re-identification.
\newblock In {\em European Conference on Computer Vision}, pages 258--274.
  Springer, 2020.

\bibitem{ristani2016MTMC}
Ergys Ristani, Francesco Solera, Roger Zou, Rita Cucchiara, and Carlo Tomasi.
\newblock Performance measures and a data set for multi-target, multi-camera
  tracking.
\newblock In {\em European Conference on Computer Vision}, pages 17--35.
  Springer, 2016.

\bibitem{ruan2020correlation}
Weijian Ruan, Chao Liang, Yi Yu, Zheng Wang, Wu Liu, Jun Chen, and Jiayi Ma.
\newblock Correlation discrepancy insight network for video re-identification.
\newblock {\em ACM Trans. Multimedia Comput. Commun. Appl.}, 16(4):1--21, 2020.

\bibitem{CVPR2015facenet}
Florian Schroff, Dmitry Kalenichenko, and James Philbin.
\newblock Facenet: A unified embedding for face recognition and clustering.
\newblock In {\em Proceedings of the IEEE/CVF Conference on Computer Vision and
  Pattern Recognition}, pages 815--823, 2015.

\bibitem{shen2019multi}
Chen Shen, Zhongming Jin, Wenqing Chu, Rongxin Jiang, Yaowu Chen, Guo-Jun Qi,
  and Xian-Sheng Hua.
\newblock Multi-level similarity perception network for person
  re-identification.
\newblock {\em ACM Trans. Multimedia Comput. Commun. Appl.}, 15(2):1--19, 2019.

\bibitem{AAAI2018QEN}
Guanglu Song, Biao Leng, Yu Liu, Congrui Hetang, and Shaofan Cai.
\newblock Region-based quality estimation network for large-scale person
  re-identification.
\newblock In {\em Proceedings of the AAAI Conference on Artificial
  Intelligence}, volume~32, 2018.

\bibitem{ICCVSubramaniam2019Co}
Arulkumar Subramaniam, Athira Nambiar, and Anurag Mittal.
\newblock Co-segmentation inspired attention networks for video-based person
  re-identification.
\newblock In {\em Proceedings of the IEEE/CVF International Conference on
  Computer Vision}, pages 562--572, 2019.

\bibitem{ECCVsun2018beyond}
Yifan Sun, Liang Zheng, Yi Yang, Qi Tian, and Shengjin Wang.
\newblock Beyond part models: Person retrieval with refined part pooling (and a
  strong convolutional baseline).
\newblock In {\em European Conference on Computer Vision}, pages 480--496,
  2018.

\bibitem{sutskever2013importance}
Ilya Sutskever, James Martens, George Dahl, and Geoffrey Hinton.
\newblock On the importance of initialization and momentum in deep learning.
\newblock In {\em International Conference on Machine Learning}, pages
  1139--1147. PMLR, 2013.

\bibitem{tang2022harmonious}
Zengming Tang and Jun Huang.
\newblock Harmonious multi-branch network for person re-identification with
  harder triplet loss.
\newblock {\em ACM Trans. Multimedia Comput. Commun. Appl.}, 18(4):1--21, 2022.

\bibitem{TIP2020cdpm}
Kan Wang, Changxing Ding, Stephen~J Maybank, and Dacheng Tao.
\newblock Cdpm: Convolutional deformable part models for semantically aligned
  person re-identification.
\newblock {\em IEEE Transactions on Image Processing}, 29:3416--3428, 2020.

\bibitem{wang2021batch}
Kan Wang, Pengfei Wang, Changxing Ding, and Dacheng Tao.
\newblock Batch coherence-driven network for part-aware person
  re-identification.
\newblock {\em IEEE Transactions on Image Processing}, 30:3405--3418, 2021.

\bibitem{ECCV2014vid}
Taiqing Wang, Shaogang Gong, Xiatian Zhu, and Shengjin Wang.
\newblock Person re-identification by video ranking.
\newblock In {\em European Conference on Computer Vision}, pages 688--703.
  Springer, 2014.

\bibitem{CVPRwang2017nonlocal}
Xiaolong Wang, Ross Girshick, Abhinav Gupta, and Kaiming He.
\newblock Non-local neural networks.
\newblock In {\em Proceedings of the IEEE/CVF Conference on Computer Vision and
  Pattern Recognition}, pages 7794--7803, 2017.

\bibitem{CVPR2018exploit}
Yu Wu, Yutian Lin, Xuanyi Dong, Yan Yan, Wanli Ouyang, and Yi Yang.
\newblock Exploit the unknown gradually: One-shot video-based person
  re-identification by stepwise learning.
\newblock In {\em Proceedings of the IEEE/CVF Conference on Computer Vision and
  Pattern Recognition}, pages 5177--5186, 2018.

\bibitem{ICCVChen2017joint}
Shuangjie Xu, Yu Cheng, Kang Gu, Yang Yang, Shiyu Chang, and Pan Zhou.
\newblock Jointly attentive spatial-temporal pooling networks for video-based
  person re-identification.
\newblock In {\em Proceedings of the IEEE International Conference on Computer
  Vision}, pages 4733--4742, 2017.

\bibitem{xu2022bire}
Sheng Xu, Chang Liu, Baochang Zhang, Jinhu L{\"u}, Guodong Guo, and David
  Doermann.
\newblock Bire-id: Binary neural network for efficient person re-id.
\newblock {\em ACM Trans. Multimedia Comput. Commun. Appl.}, 18(1s):1--22,
  2022.

\bibitem{CVPR2020MGH}
Yichao Yan, Jie Qin, Jiaxin Chen, Li Liu, Fan Zhu, Ying Tai, and Ling Shao.
\newblock Learning multi-granular hypergraphs for video-based person
  re-identification.
\newblock In {\em Proceedings of the IEEE/CVF Conference on Computer Vision and
  Pattern Recognition}, pages 2899--2908, 2020.

\bibitem{CVPR2020STG}
Jinrui Yang, Wei-Shi Zheng, Qize Yang, Yingcong Chen, and Qi Tian.
\newblock Spatial-temporal graph convolutional network for video-based person
  re-identification.
\newblock In {\em Proceedings of the IEEE/CVF Conference on Computer Vision and
  Pattern Recognition}, pages 3289--3299, 2020.

\bibitem{yang2017enhancing}
Xun Yang, Meng Wang, Richang Hong, Qi Tian, and Yong Rui.
\newblock Enhancing person re-identification in a self-trained subspace.
\newblock {\em ACM Trans. Multimedia Comput. Commun. Appl.}, 13(3):1--23, 2017.

\bibitem{zhang2021hat}
Guowen Zhang, Pingping Zhang, Jinqing Qi, and Huchuan Lu.
\newblock Hat: Hierarchical aggregation transformers for person
  re-identification.
\newblock In {\em Proceedings of the 29th ACM International Conference on
  Multimedia}, pages 516--525, 2021.

\bibitem{zhang2021pixel}
Wenyu Zhang, Qing Ding, Jian Hu, Yi Ma, and Mingzhe Lu.
\newblock Pixel-wise graph attention networks for person re-identification.
\newblock In {\em Proceedings of the 29th ACM International Conference on
  Multimedia}, pages 5231--5238, 2021.

\bibitem{CVPR2020MGRAFA}
Zhizheng Zhang, Cuiling Lan, Wenjun Zeng, and Zhibo Chen.
\newblock Multi-granularity reference-aided attentive feature aggregation for
  video-based person re-identification.
\newblock In {\em Proceedings of the IEEE/CVF Conference on Computer Vision and
  Pattern Recognition}, pages 10407--10416, 2020.

\bibitem{CVPR2019attribute}
Yiru Zhao, Xu Shen, Zhongming Jin, Hongtao Lu, and Xian-sheng Hua.
\newblock Attribute-driven feature disentangling and temporal aggregation for
  video person re-identification.
\newblock In {\em Proceedings of the IEEE/CVF Conference on Computer Vision and
  Pattern Recognition}, pages 4913--4922, 2019.

\bibitem{ECCV2016mars}
Liang Zheng, Zhi Bie, Yifan Sun, Jingdong Wang, Chi Su, Shengjin Wang, and Qi
  Tian.
\newblock Mars: A video benchmark for large-scale person re-identification.
\newblock In {\em European Conference on Computer Vision}, pages 868--884.
  Springer, 2016.

\bibitem{zheng2015scalable}
Liang Zheng, Liyue Shen, Lu Tian, Shengjin Wang, Jingdong Wang, and Qi Tian.
\newblock Scalable person re-identification: A benchmark.
\newblock In {\em Proceedings of the IEEE International Conference on Computer
  Vision}, pages 1116--1124, 2015.

\bibitem{CVPRzheng2019re}
Meng Zheng, Srikrishna Karanam, Ziyan Wu, and Richard~J Radke.
\newblock Re-identification with consistent attentive siamese networks.
\newblock In {\em Proceedings of the IEEE/CVF Conference on Computer Vision and
  Pattern Recognition}, pages 5735--5744, 2019.

\bibitem{zheng2017discriminatively}
Zhedong Zheng, Liang Zheng, and Yi Yang.
\newblock A discriminatively learned cnn embedding for person reidentification.
\newblock {\em ACM Trans. Multimedia Comput. Commun. Appl.}, 14(1):1--20, 2017.

\bibitem{zhong2021glance}
Xubin Zhong, Xian Qu, Changxing Ding, and Dacheng Tao.
\newblock Glance and gaze: Inferring action-aware points for one-stage
  human-object interaction detection.
\newblock In {\em Proceedings of the IEEE/CVF Conference on Computer Vision and
  Pattern Recognition}, pages 13234--13243, 2021.

\bibitem{zhong2017random}
Zhun Zhong, Liang Zheng, Guoliang Kang, Shaozi Li, and Yi Yang.
\newblock Random erasing data augmentation.
\newblock In {\em Proceedings of the AAAI Conference on Artificial
  Intelligence}, volume~34, pages 13001--13008, 2020.

\bibitem{CVPR2017see}
Zhen Zhou, Yan Huang, Wei Wang, Liang Wang, and Tieniu Tan.
\newblock See the forest for the trees: Joint spatial and temporal recurrent
  neural networks for video-based person re-identification.
\newblock In {\em Proceedings of the IEEE/CVF Conference on Computer Vision and
  Pattern Recognition}, pages 4747--4756, 2017.

\end{thebibliography}
}

\clearpage

\end{document}